\documentclass[final]{cvpr}

\usepackage{times}
\usepackage{epsfig}
\usepackage{graphicx}
\usepackage{amsmath}
\usepackage{amssymb}

\usepackage{appendix}

\usepackage{multirow}
\usepackage{booktabs}
\usepackage{subfig}

\newcommand{\tablestyle}[2]{\setlength{\tabcolsep}{#1}\renewcommand{\arraystretch}{#2}\centering\small}

\newlength\savewidth\newcommand\shline{\noalign{\global\savewidth\arrayrulewidth
  \global\arrayrulewidth 1pt}\hline\noalign{\global\arrayrulewidth\savewidth}}

\usepackage[pagebackref=true,breaklinks=true,colorlinks,bookmarks=false]{hyperref}

\def\confYear{CVPR 2021}

\begin{document}

\title{Self-supervised Motion Learning from Static Images}

\author{Ziyuan Huang$^{1,2}$, Shiwei Zhang$^2$, Jianwen Jiang$^2$, Mingqian Tang$^2$, Rong Jin$^2$, Marcelo H. Ang Jr$^1$\\
$^1$ National University of Singapore, Singapore\\
$^2$ Alibaba Group, China\\
{\tt\small ziyuan.huang@u.nus.edu, mpeangh@nus.edu.sg}\\
{\tt\small \{zhangjin.zsw, jianwen.jjw, mingqian.tmq,jinrong.jr\}@alibaba-inc.com}
}

\maketitle
\pagestyle{empty}  
\thispagestyle{empty} 

\begin{abstract}
    Motions are reflected in videos as the movement of pixels, and actions are essentially patterns of inconsistent motions between the foreground and the background. 
    To well distinguish the actions, especially those with complicated spatio-temporal interactions, correctly locating the prominent motion areas is of crucial importance. 
    However, most motion information in existing videos are difficult to label and training a model with good motion representations with supervision will thus require a large amount of human labour for annotation.
    In this paper, we address this problem by self-supervised learning. 
    Specifically, we propose to learn \textbf{Mo}tion from \textbf{S}tatic \textbf{I}mages (MoSI). The model learns to encode motion information by classifying pseudo motions generated by MoSI. 
    We furthermore introduce a static mask in pseudo motions to create local motion patterns, 
    which forces the model to additionally locate notable motion areas for the correct classification.
    We demonstrate that MoSI can discover regions with large motion even without fine-tuning on the downstream datasets. 
    As a result, the learned motion representations boost the performance of tasks requiring understanding of complex scenes and motions, \textit{i.e.}, action recognition.
    Extensive experiments show the consistent and transferable improvements achieved by MoSI.
    Codes will be soon released.

\end{abstract}

\section{Introduction}
\begin{figure}[t]
\begin{center}
   \includegraphics[width=0.95\linewidth]{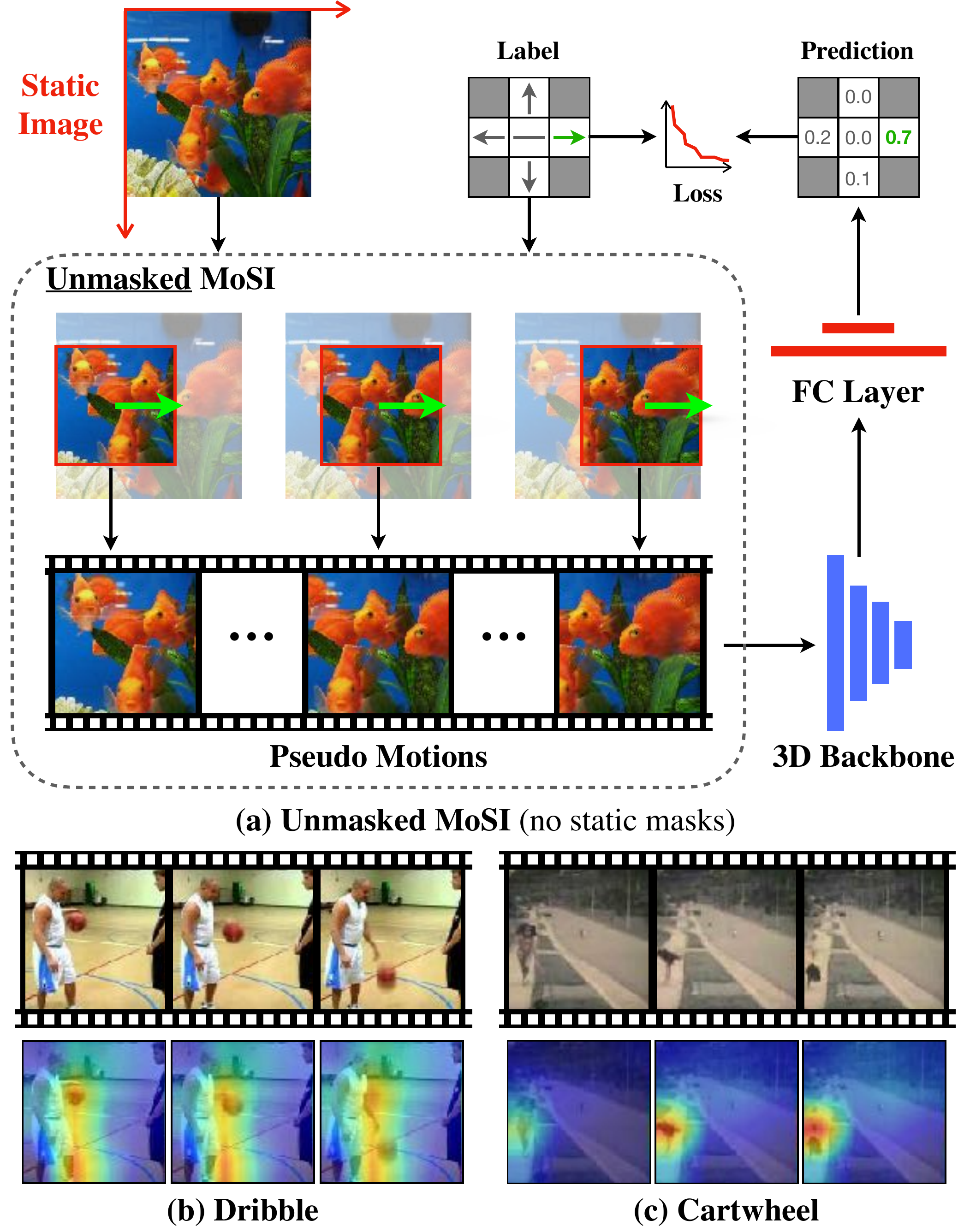}
\end{center}\vspace{-5.5mm}
   \caption{(a) \textbf{Unmasked MoSI} constructs image sequences with pseudo motions from static images. The model is trained to encode motions by predicting the direction and speed of the pseudo motions. For simplicity, the speed granularity here is set as $K=1$ (see Sec.~\ref{sec:pseudo-motions}). 
   (b), (c) Grad-CAM~\cite{gradcam} visualizations on HMDB51 videos for the conv5 pre-trained by our MoSI on ImageNet~\cite{imagenet}, where the model locates prominent motions even without fine-tuning on the downstream dataset (See more in Sec.~\ref{sec:understanding}).
  }\vspace{-5mm}
\label{fig:unmasked_mosi}
\end{figure}

Understanding motion patterns is a key challenge in many video understanding problems such as action recognition~\cite{slowfast}, action localization~\cite{CDC} and action detection~\cite{SSN}. 
A suitable way to encode motions can significantly boost the performance in those tasks~\cite{twostream}.
Early works represent motions using hand-crafted features~\cite{stackedfishervectors,improvedtrajectories,BoVW} based on dense trajectories~\cite{densetrajectories} and optical flow~\cite{opticalflow}. 
With the successful application of deep neural networks~\cite{resnet,alexnet,largescalevideo} and the construction of large scale image and video datasets~\cite{imagenet,kinetics400}, endeavors have been made to design architectures to extract meaningful motion features~\cite{stm,slowfast,nonlocal,corrnet,trajectoryconv,twostream}. 
Despite their powerful capability of modeling dynamic variations between frames, the 3D convolutional models require a large amount of manually labeled videos to achieve a good generalized performance~\cite{retrace}.

Recently, self-supervised learning has emerged as a powerful technique for training the model without labeled data in both image and video paradigm~\cite{vcop,dpc,ST-puzzle,contextprediction,inpainting}. 
These methods learn visual representations by exploiting inherent structures of the unlabeled images or videos, for instance, by predicting the correct order of spatial or temporal sequences~\cite{vcop,contextprediction,oddoneout,opn} or by predicting partial contents~\cite{dpc,memdpc,inpainting}. 
Because videos naturally have an extra axis of time compared to images, some methods manipulate the temporal dimension and predict the playback speeds~\cite{speednet,prp}. 
Although some of the efforts were able to capture the motion information implicitly, almost none of them aims to model motion information of videos explicitly in a self-supervised fashion. 

In this work, we seek to train the video model to directly distinguish different motion patterns. 
The objective is for the model to encode meaningful motion information, so that prominent motions can be discovered and attended to during fine-tuning. 
Since directly generating predefined motion patterns from a video set may be difficult, we leverage static images for motion generation. 
Formally, we propose a learning framework that learns motions directly from images (MoSI). Its general structure is shown in Fig.~\ref{fig:unmasked_mosi}. 
Given the desired direction and the speed of the motions, MoSI generates pseudo motions from static images. 
By correctly classifying the direction and speed of the movement in the image sequence, models trained with MoSI is able to well encode motion patterns. 
Furthermore, a static mask (Fig.~\ref{fig:static_mask}) is applied to the pseudo motion sequences. 
This produces inconsistent motions between the masked area and the unmasked one, which guides the network to focus on the inconsistent local motions. 
We term the one with and without static masks as MoSI and unmasked MoSI respectively. 
Conceptually, the idea of masked MoSI is closely related to attention learning, where the network learns to attend more to the moving areas in videos explicitly created by pseudo motion. 
Different from the attention mechanism~\cite{senet,sknet,tea}, where attention is generated by carefully designed architectures, the attention learned by MoSI is achieved by purely altering the training data.

To the best of our knowledge, this is the first time that static images are used as the data source for pre-training video models. 
Using MoSI, we are able to exploit large-scale image datasets such as ImageNet~\cite{imagenet} to train video models.
Although images contain less information about dynamics that are intrinsic in videos, the representations learned with MoSI can be as powerful as those learned using videos in terms of motion understanding.
Extensive empirical studies with HMDB51 and UCF101 further demonstrate the effectiveness of MoSI. Compared with other previously published works, we show that the proposed MoSI reaches new state-of-the-art results for learning video representations using RGB modality. 

\section{Related Work}
\noindent\textbf{Motion learning by architectures.} 
Motion information are crucial for understanding videos. 
There are mainly two popular architectures that are frequently used for extracting video features, respectively two-stream networks~\cite{twostream,twostreamfusion,tsn,lgd,mars} and 3D convolutional networks~\cite{i3d,retrace,r21d,p3d,csn}. 
Two-stream networks extracts motions representations explicitly from optical flows, while 3D structures apply convolutions on the temporal dimension~\cite{p3d,r21d} or space-time cubics~\cite{i3d,csn,retrace} to extract motion cues implicitly. 
Besides these two architectures, different motion encodings are proposed to better handle motions in videos~\cite{tea,stm,corrnet}.
Compared to complicated structure designs that aim at better representing motions, our MoSI can take any video models as the backbone. For simplicity, structures proposed in \cite{r21d} are adopted for our experiments.

\noindent\textbf{Self-supervised image representation learning.} 
Self-supervised learning is proven to be a powerful tool for learning representations that are useful to down-stream tasks without requiring labeled data. 
Using image as data sources, there are patch-based approaches~\cite{contextprediction,jigsaw,rrm,jigsaw++} that are inspired by natural language processing methods~\cite{wordrepresentations}, and image-level pretext tasks, such as image inpainting~\cite{inpainting}, image colorization~\cite{colorization}, motion segment prediction~\cite{motionmask} and predicting image rotations~\cite{rotnet}. 
The most similar to our work is \cite{motionmask}, where labels are generated by grouping pixels that share the same movement together in videos. 
There are two crucial differences: (a) the aim of \cite{motionmask} is to learn pixels that belong to the same object by motion segmentation, while our MoSI is proposed to learn motion cues for understanding videos; (b) \cite{motionmask} exploits videos to learn image representations, while MoSI takes advantage of images to learn video representations.

\noindent\textbf{Self-supervised video representation learning.} 
With an extra time dimension, videos provides rich static and dynamic information, and there is thus an abundant supply of various supervision signals. 
A natural way is to extend patch-based context prediction to spatio-temporal scenarios, such as spatio-temporal puzzles~\cite{ST-puzzle}, video cloze procedure~\cite{vcp} and frame/clip order prediction~\cite{opn,vcop,oddoneout}. 
Besides the extension of image based supervisions, recent works propose to learn representations by predicting future frames~\cite{dpc,memdpc}. 
In addition, supervision signals can be generated by purely manipulating the time axis. 
Representative works include speed up prediction~\cite{speednet} and play back rate prediction~\cite{prp}. 
All previous video representation learning methods exploit videos as the data source. 
Hence, the motion patterns have not yet been able to be explicitly learned due to the difficulty of generating predefined motion patterns from videos. 
In this work, we take images as our data source, and generate deterministic motion patterns for directly learning motion representations. 

\section{Motion Learning from Static Images}
The goal of MoSI is to learn motion representations. Because directly generating predefined motions from videos could be difficult, MoSI exploit images to generate samples for motion learning. 
Specifically, MoSI generates pseudo motions with different speeds and directions. 
To correctly predict the motion pattern, the 3D video backbone is required to distinguish different motion patterns.
In addition, to mimic the actions in actual videos, where there exist inconsistent motions between the foreground and the background, we apply a static mask to the generated pseudo motions. 
In this way, the network is additionally required to locate prominent motion areas and attend less to the background. 
In short, there are two core components in the proposed MoSI, respectively the pseudo motions and the static masks, which will be discussed in Sec.~\ref{equ:pseudo-motion} and Sec.~\ref{sec:static-masks} respectively. 
In the following sections, we refer to the framework as MoSI and unmasked MoSI respectively for the variants with and without static masks. 
\subsection{Pseudo Motions}
\label{sec:pseudo-motions}
The first component is the pseudo motions. The generation process is visualized in Fig.~\ref{fig:unmasked_mosi}. 
Given the motion label $(x,y)$ sampled from the label pool $\mathbb{L}$, MoSI crops an continuous sequence of images $\mathbf{u}\in\mathbb{R}^{N\times L\times L}$ from the input image (which we term as source image). 
$N$ and $L$ are selected in accordance to the number of frames and crop size in the downstream task. 
The generated pseudo motion sequence is then used as the input to the video backbone for motion classification.

\noindent\textbf{Label pool. }
The motion patterns generated by MoSI consists of two axes, respectively a horizontal axis and a vertical axis. 
The positive direction for them are respectively toward right and down, as in Fig.~\ref{fig:unmasked_mosi}. 
For each axis, there are $C=2\times K + 1$ speeds, where $K$ denotes the granularity of the speeds in one direction (\textit{e.g.}, the positive direction on the horizontal axis). 
This corresponds to the motion speed set $\mathbb{S}=\{-K, ..., -1, 0, 1, ... K\}$, where negative values indicate moving in the negative direction of the corresponding axis.
$K$ is set to be larger than 1, since we want the network to learn not only the existence but also the magnitude of motions. 
For simplicity, we decouple the motions for two axis, which means for each label, a non-zero speed only exists on one axis. 
Therefore, the total size of the label pool is $C_T = 2\times C-1 = 4\times K + 1$ with $K$ labels for each direction and $1$ label denoting static sequence. 
The label pool can be expressed as follows for each label index $i$:
\begin{equation}
    \mathbb{L}=\{i: (x,y)|x\in \mathbb{S}, y\in \mathbb{S}, xy=0 \} \ .
\end{equation}
\noindent Note: It is crucial to generate motions for both axes, because the motion patterns in videos can be both horizontal and vertical. 
See Sec.~\ref{sec:understanding} for the empirical results. 

\noindent\textbf{Pseudo motion generation.} 
To generate the samples with different speeds, we define the moving distance from the start to the end of the pseudo motion sequences. 
For source image with the size of $H\times W$, the distance $\overrightarrow{D}=(D_x, D_y)$ for the pseudo motion of speed $(x,y)\in \mathbb{L}$ is defined as:
\begin{equation}
\left\{
    \begin{aligned}
        D_x &= \frac{(W-L)x}{K}, \ \text{if} \ x\neq 0 \ \text{else} \ D_x=0 \\
        D_y &= \frac{(H-L)y}{K}, \ \text{if} \ y\neq 0 \  \text{else} \ D_y=0  
    \end{aligned}\right. \ .
\label{equ:pseudo-motion}
\end{equation}
\noindent Note that the value of $D_x$ and $D_y$ could be negative, which denotes moving in the negative direction of an axis.

\begin{figure}[t]
\begin{center}
  \includegraphics[width=\linewidth]{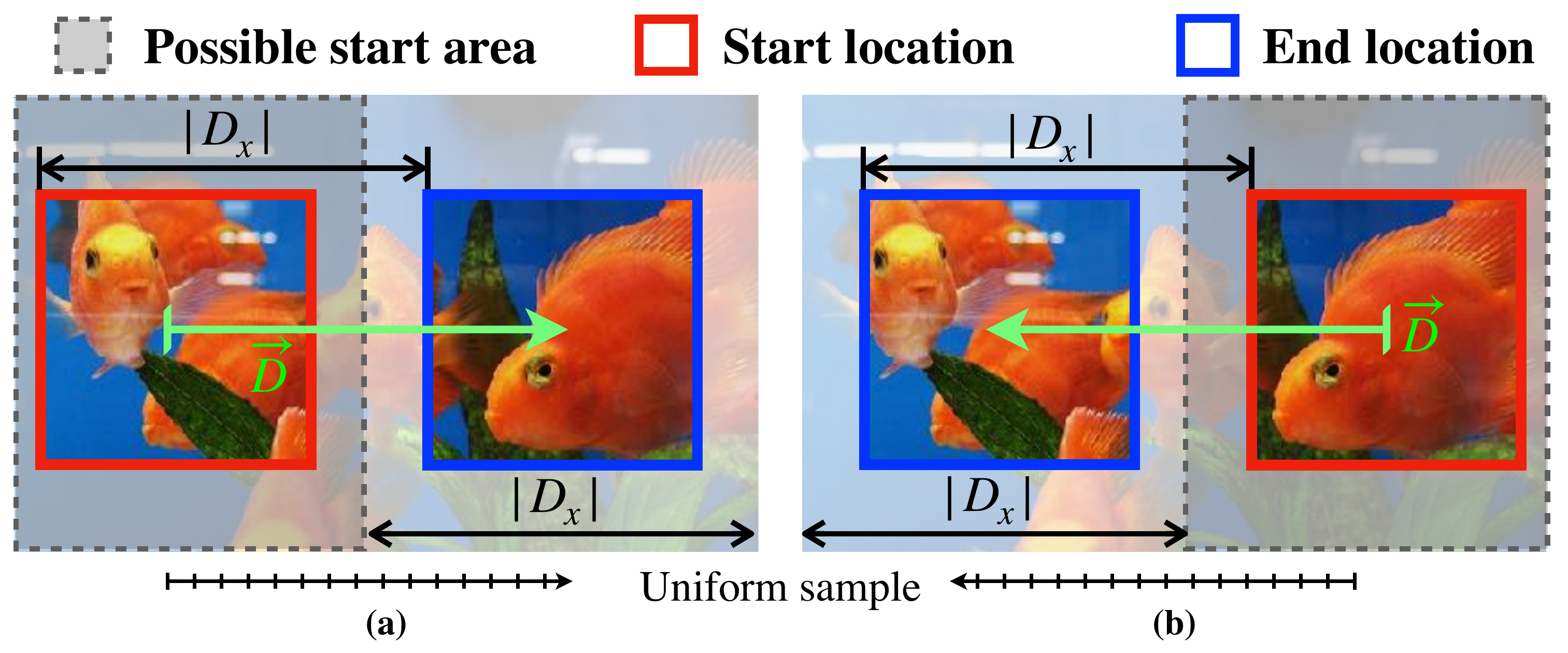}
\end{center}\vspace{-3mm}
  \caption{\textbf{Exemplar pseudo motion generation processes.} (a) and (b) generates motions respectively in the positive and negative direction of the horizontal axis. The start image is sampled randomly from the \textit{possible start area} and the position of the end image is calculated using the distance $D_x$ and $D_y$. The \textit{possible start area} is determined so that the end image is located within the source image. $N$ frames are uniformly sampled between the start and the end positions from the source image. 
  }
\label{fig:generation}
\end{figure}

The start location $\overrightarrow{l_{start}} = (x_{start}, y_{start})$ is randomly sampled from a certain area which ensures the end location $\overrightarrow{l_{end}}=\overrightarrow{l_{start}}+\overrightarrow{D}$ is located completely within the source image, as demonstrated in Fig.~\ref{fig:generation}. 
For example, if $D_x>0$, the distance between the right border of both the start image and the source image should be at least $|D_x|$.
For label $(x,y)=(0,0)$, where the sampled image sequence is static on both axis, the start location is selected from the whole image with uniform distribution.
$N$ images are then sampled with uniform gaps from the source image between the start position $\overrightarrow{l_{start}}$ and the end position $\overrightarrow{l_{end}}$.

\noindent\textbf{Classification.} The generated image sequence $\mathbf{u}$ is then fed into a 3D backbone network and a linear classifier. 
Following \cite{speednet,rotnet}, we employ the same-batch training technique, where each batch contains all transformed image sequences generated from the same source image. 
This means for each source image, $C_T$ image sequences of pseudo motions are generated and included in the same mini-batch. 
This is found to be significantly effective for reducing the artificial cues. 
The model is trained by cross entropy loss.
\begin{figure}[t]
\begin{center}
  \includegraphics[width=1\linewidth]{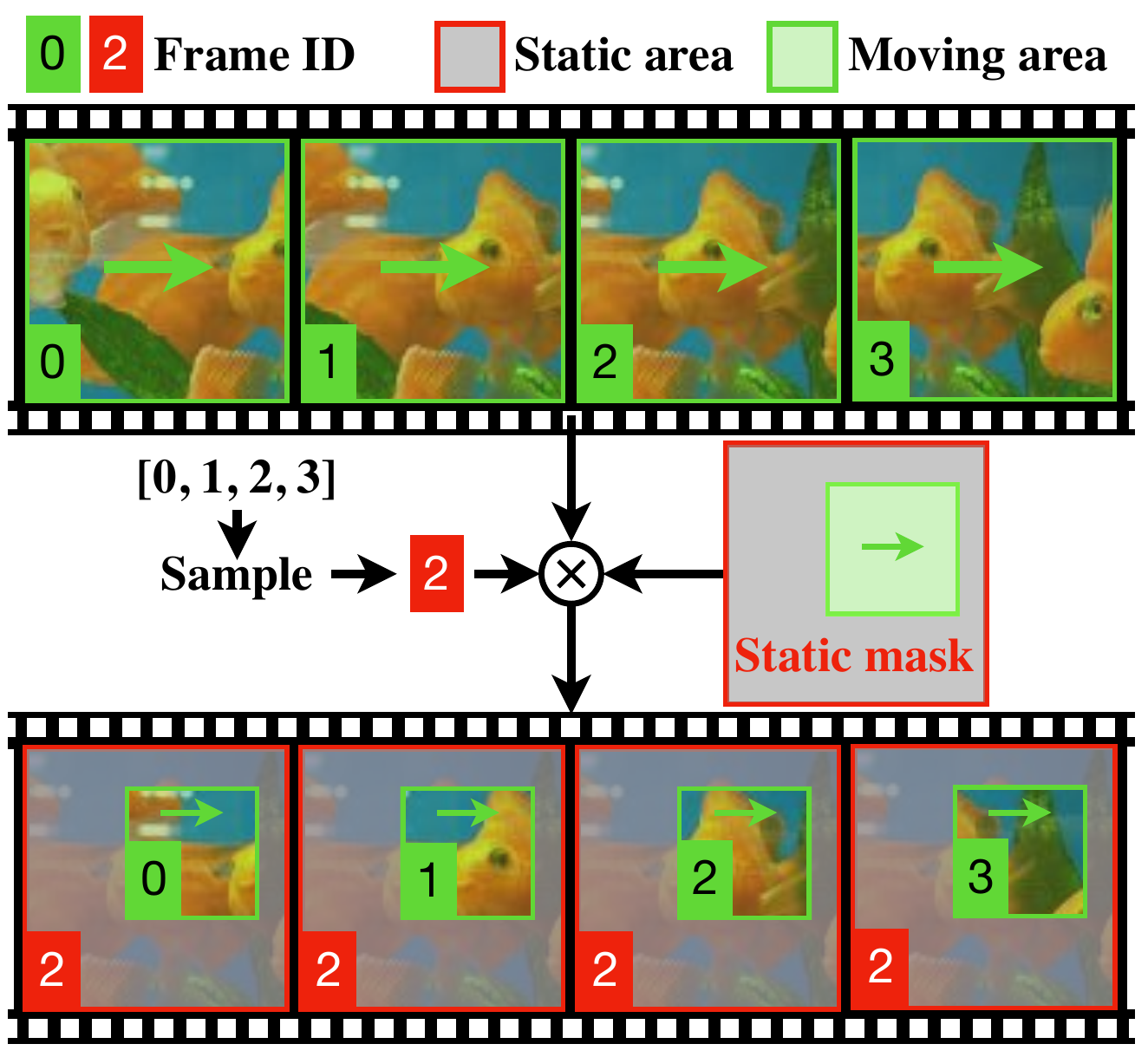}
\end{center}\vspace{-3mm}
  \caption{\textbf{Exemplar static mask applied on the sampled image sequence.} One of the images in the generated pseudo motions is selected to replace the contents of each image in the static area, while the contents in the moving area are not altered. In this case, the third image (ID=2) in the sequence is selected as the background. The green arrows indicate the moving direction of the bounding boxes in the source image. Essentially, the static mask is a filter that only allows contents in a certain area to be kept the same as the input. The unmasked variant (upper sequence) represents global motion. The static masks creates local motion patterns that are inconsistent with the background. 
  }
\label{fig:static_mask}
\end{figure}
\subsection{Static Masks}
\label{sec:static-masks}
By correctly classifying pseudo motions with different directions and magnitudes, the model is able to recognize different motion patterns. 
However, since for most videos, actions occur in a constrained area rather than all the spatial locations, one is expected to recognize not only global motion patterns, but also inconsistent motions between the foreground and the background.
Another drawback for the model to understand global motion is that the model will possibly focus on just several pixels, as all the motion patterns (speed and direction) in the image sequences are essentially the same. 
This creates an obvious artificial cue~\cite{contextprediction,dpc,arrowoftime} that hinders the true capability of the model to understand motions. 
To this end, we introduce static masks as the second core component of the proposed MoSI. 

Static masks divide the spatial location into two groups, respectively masked area and unmasked area, as in Fig.~\ref{fig:static_mask}. 
The masked area is regarded as the background and the motions within this area is thus removed, by setting the content of this area in all images in $\mathbf{u}$ to the $q$-th image $\mathbf{u}_q$. 
On the other hand, the original contents (\textit{i.e.}, the motions) are retained in the unmasked area of the image sequence $\mathbf{u}$. 
For simplicity, the unmasked area is by default a square area within the image, with the size of $L_m\times L_m$. 
Formally, given the masked pixels $\mathbf{m}$, the content of the $p$-th image is determined by:
\begin{equation}
    \tilde{\mathbf{u}}_p = M(\mathbf{u}_p, \mathbf{m})=\left\{
    \begin{aligned}
        &\mathbf{u}_q, \ \text{if}\  (a,b)\in \mathbf{m}\\
        &\mathbf{u}_p, \ \text{if}\  (a,b)\notin \mathbf{m}
    \end{aligned}
    \right. \ ,
\end{equation}
\noindent where $(a,b)$ is the spatial location of a certain pixel, and $q$ is the randomly selected static image. 

By applying the static mask, the background area of the image sequence becomes static and the foreground is moving according to the label. 
To perform correct classification, the model is now required not only to recognize motion patterns, but also to spot where the motion is happening. 
This benefits a lot for downstream tasks such as action recognition, as the model is equipped with knowledge on where to focus even before fine-tuning on the downstream datasets. 

\begin{figure*}[t]
\begin{center}
   \includegraphics[width=1\textwidth]{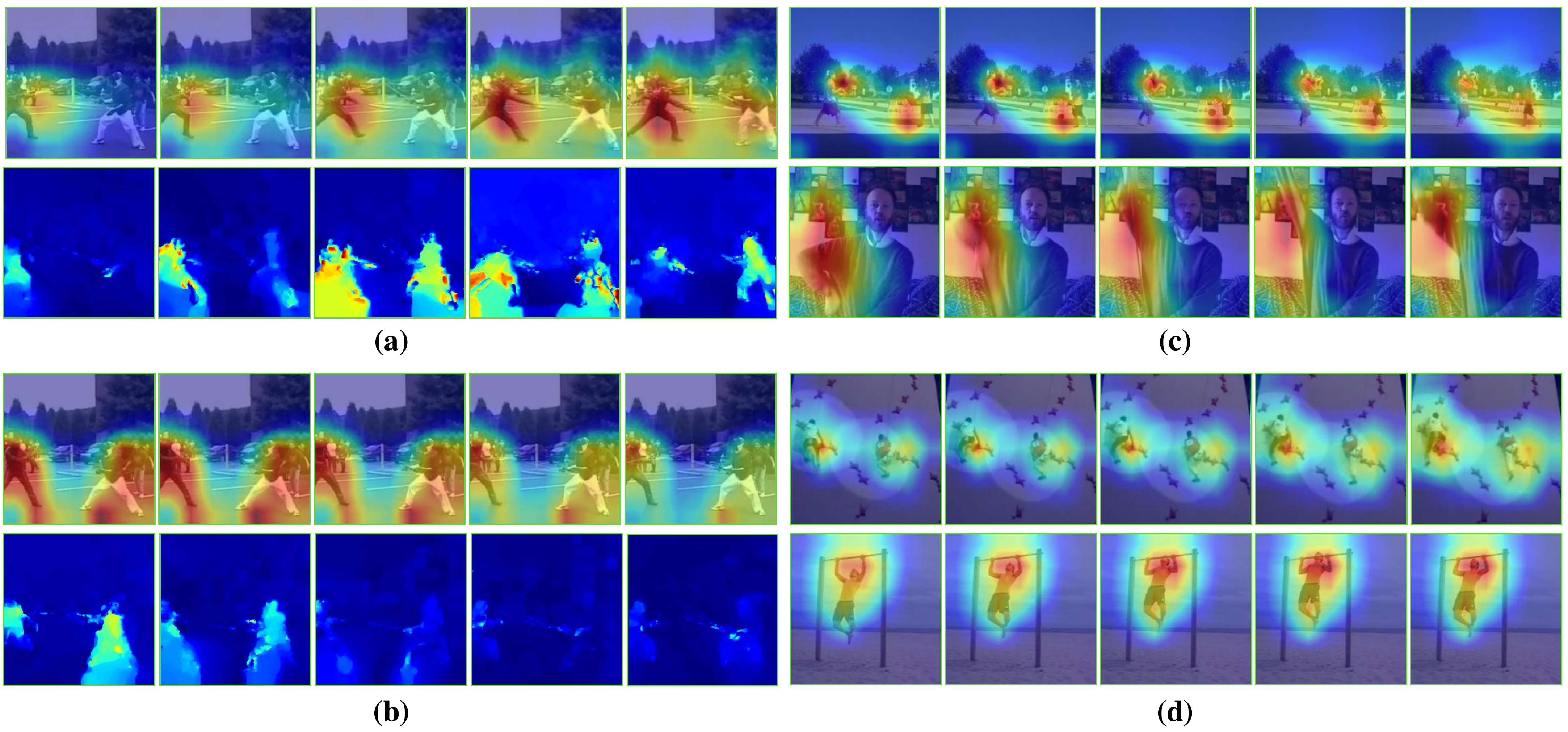}
\end{center}\vspace{-3mm}
   \caption{\textbf{Grad-CAM visualization}~\cite{gradcam} of the conv-5 layer on HMDB51 over the models trained by MoSI on ImageNet ((a), (c)) and HMDB51 ((b),(d)) \textbf{without fine-tuning}. 
   Red and blue areas denotes respectively highly and less activated areas. Although no semantic meaning has been taught through MoSI, models trained by MoSI already possess the ability to locate prominent areas according to the motions. In (a) and (b), we additionally compare with the value of optical flow calculated with $\sqrt{x^2+y^2}$. It can be observed that most highly activated areas corresponds to regions where motions are significant. 
   }
\label{fig:gradcam}
\end{figure*}

\subsection{Instantiation}

\noindent\textbf{Data preparations. }
One advantage of the proposed MoSI is that it can train video models on both video datasets and image datasets. 
This allows for exploiting a large amount of existing image-based datasets. 
For video and image datasets, the only difference is that the source images need to be first sampled from the videos in the video datasets, while for image datasets, no frame-sampling step is required. 
Specifically, for video datasets, one frame out of each video is randomly sampled as the source image. 
Using the same-batch training technique, each image generates $C_T$ samples with different labels. 
We alter the sampled source frame index for different epochs for a larger variety of visual contents. 
After obtaining the source images, we resize the source image so that the length of the short side is $L_s$. 
An $L_s\times L_s$ square area is randomly cropped from the resized image, which means $H=W=L_s$ in Eq.~\ref{equ:pseudo-motion}. 
This ensures that the motion magnitude for the same speed on both axis are the same. 

\noindent\textbf{Augmentations.}
It is shown in previous self-supervised approaches that the model tend to learn some artificial cues or trivial solutions~\cite{contextprediction,dpc,arrowoftime} that disrupts the learning of the designed objectives. 
In our case, we have introduced the static mask to avoid one possible trivial solution, where the model only need to recognize motions in an extremely small area for the prediction of the correct motion class. 
Based on that, we further randomize the location and the size of the unmasked area. 
In addition, we randomize the selection of the background frames in the MoSI. 
In Sec.~\ref{sec:understanding}, we closely investigate the effect of mask sizes and demonstrate the benefit of our randomization. 

\section{Experiments}
\noindent\textbf{Datasets and backbone.}
For pre-training with MoSI, we employ three video datasets: UCF101~\cite{ucf101}, HMDB51~\cite{hmdb}, Kinetics~\cite{kinetics400}, as well as the image dataset ImageNet~\cite{imagenet}. 
For evaluation of the learned representation, we use UCF101 and HMDB51. 
We use R(2+1)D~\cite{r21d} with 10 layers as well as R-2D3D with 18 layers as our backbone, following the configurations in \cite{vcop,dpc,memdpc,prp}.

\noindent\textbf{Self-supervised pre-training.} 
For self-supervised pre-training, we set $L_s=320$ and resize the source image to $320\times 320$ by default. 
Image sequences of length $16$ and size $112\times112$ with pseudo motions are generated from each source image and fed to the model. 
The number of speed on each axis is set to $5$, which is the minimal number for each direction to have distinct speeds. 
The total size of our label pool is thus $9$.
The side length of the unmasked area in our static mask $L_m$ is randomly sampled from $[0.3, 0.5]\times 112$. 

\noindent\textbf{Supervised action classification.} 
During supervised training on UCF101 and HMDB51 for action classification, we train the network with a batch size of 128 samples per GPU for 8 GPUs using Adam with a base learning rate of 0.002 for 300 epochs. 
For evaluation, we follow the standard protocal~\cite{vcop,r21d} using 10 clips to produce the final results and report the results on split 1 on both UCF101 and HMDB51.
Further details on both self-supervised and supervised training can be referred to the supplemental material.


\subsection{Understanding MoSI.}
\label{sec:understanding}
In this section, we investigate the models trained by MoSI. 
We use {R-2D3D} with 18 layers in this section with the same structure as in \cite{dpc}. 
The datasets used for pre-training and fine-tuning are the same unless otherwise specified. 
For each ablation experiment, only the inspected factor is altered and the rest of the settings are kept according to the ones described before. 

\noindent\textbf{What has the network learned?} We first establish some intuitive understanding of the method, by addressing the question of \textit{what has the model learned through MoSI}. 
The Grad-CAM~\cite{gradcam} visualization of the last layer in the pre-trained model is shown in Fig.~\ref{fig:gradcam}. 
Note that no fine-tuning is performed at this stage. 
As can be seen, the model has learned to pick up salient motion regions in the videos.
Especially compared to optical flow, the model trained by MoSI highlights the region where the values of the optical flow is large. 
Furthermore, despite the model is only given pseudo motions as training data, it is able to transfer the knowledge onto real videos with more complex spatio-temporal relations to discover locate areas with a large motion across different frames. 
In addition, the prior square motion area does not constrain the model to only look for square areas with motions. 
Surprisingly, given only prior knowledge of one possible moving region, the model learns to generalize to find multiple prominent motion areas. 

\begin{figure}[t]
\begin{center}
   \includegraphics[width=1\linewidth]{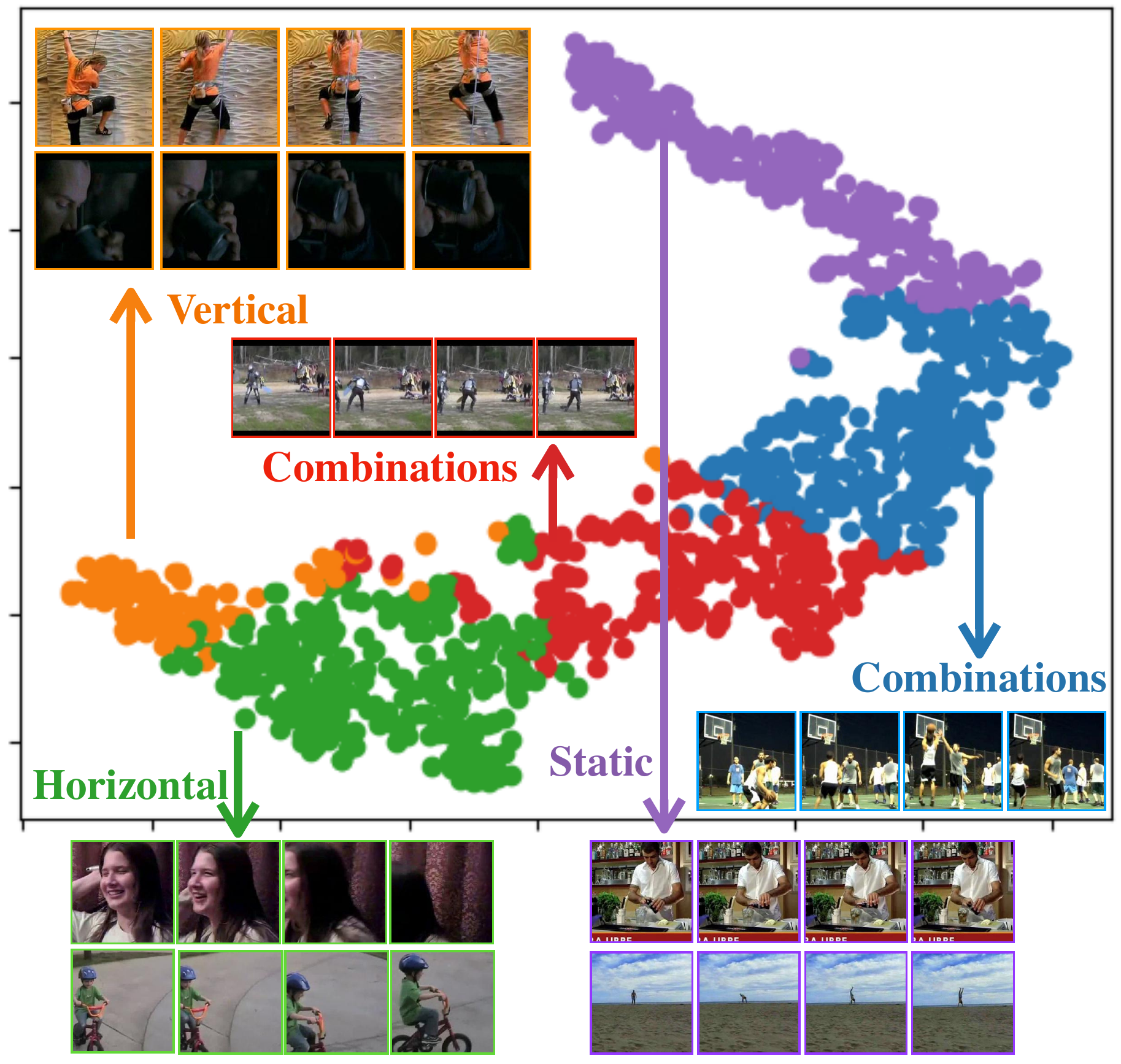}
\end{center}\vspace{-5mm}
   \caption{\textbf{T-SNE visualization of clustering results on HMDB51.} The representations are generated by unmasked MoSI for easy analysis, and without fine-tuning to directly evaluate the learned representation.
   }
\label{fig:clustering}
\end{figure}

We further cluster the learned representation on HMDB51, as in Fig.~\ref{fig:clustering}.
Although video data are naturally heterogeneous and consist of complex combinations of motions, we can still observe that the movements of a large portions of pixels in three of the five clusters are easily observable, which are vertical, horizontal and static. 
The motions in the other two clusters are hard to be uniformly described as the multiple movements are present at the same time.

\noindent\textbf{Baseline comparison. }
We then fine-tune the pre-trained model on the action classification task. 
As in Table~\ref{tab:baseline}, models trained with MoSI achieve notable improvements on both datasets. 
The improvement on HMDB51 reaches 16.5\% when using both axes. 
The performances with only one axis are weaker compared to two axes, but they still outperform the baseline by a notable margin.
On UCF101, the improvement of MoSI reaches 7.3\%. 
However, the benefit of using two axes is smaller. 
This is partially because the motion cue can be of less importance in classifying UCF101 videos, where even using only one image could achieve satisfactory classification performance, as shown in~\cite{finegym}. 

\begin{table}[t]
    \centering
\tablestyle{5pt}{1.0}
\footnotesize
\begin{tabular}{c|c|cc|c}
 \small Dataset & \small Initialization & \small Label-X & \small Label-Y & \small Top1-Acc\\
\shline
 \multirow{4}{*}{UCF101} & From scratch & - & - & 64.5  \\
 \cline{2-5}
 ~ & \multirow{3}{*}{\small MoSI} & \checkmark & \checkmark & \textbf{71.8} \scriptsize{(+7.3)}\\
 ~ & ~ & \checkmark & $\times$ & 71.6 \scriptsize{(+7.1)} \\
 ~ & ~ & $\times$ & \checkmark & 69.9 \scriptsize{(+5.4)}\\
 \hline
 \multirow{4}{*}{HMDB51} & From scratch & - & - & 30.4  \\
 \cline{2-5}
~ & \multirow{3}{*}{\small MoSI} & \checkmark & \checkmark & \textbf{47.0} \scriptsize{(+16.6)}\\
 ~ & ~ & \checkmark & $\times$ & 44.9 \scriptsize{(+14.5)} \\
 ~ & ~ & $\times$ & \checkmark & 43.1 \scriptsize{(+12.7)}\\
 
\end{tabular}\vspace{-3mm}
    \caption{\textbf{Baseline comparison} of models trained using MoSI. Models pretrained with MoSI achieves a notable improvement over the baselines that are trained from scratch. }
    \label{tab:baseline}
\end{table}

\begin{figure}[t]
\begin{center}
   \includegraphics[width=1\linewidth]{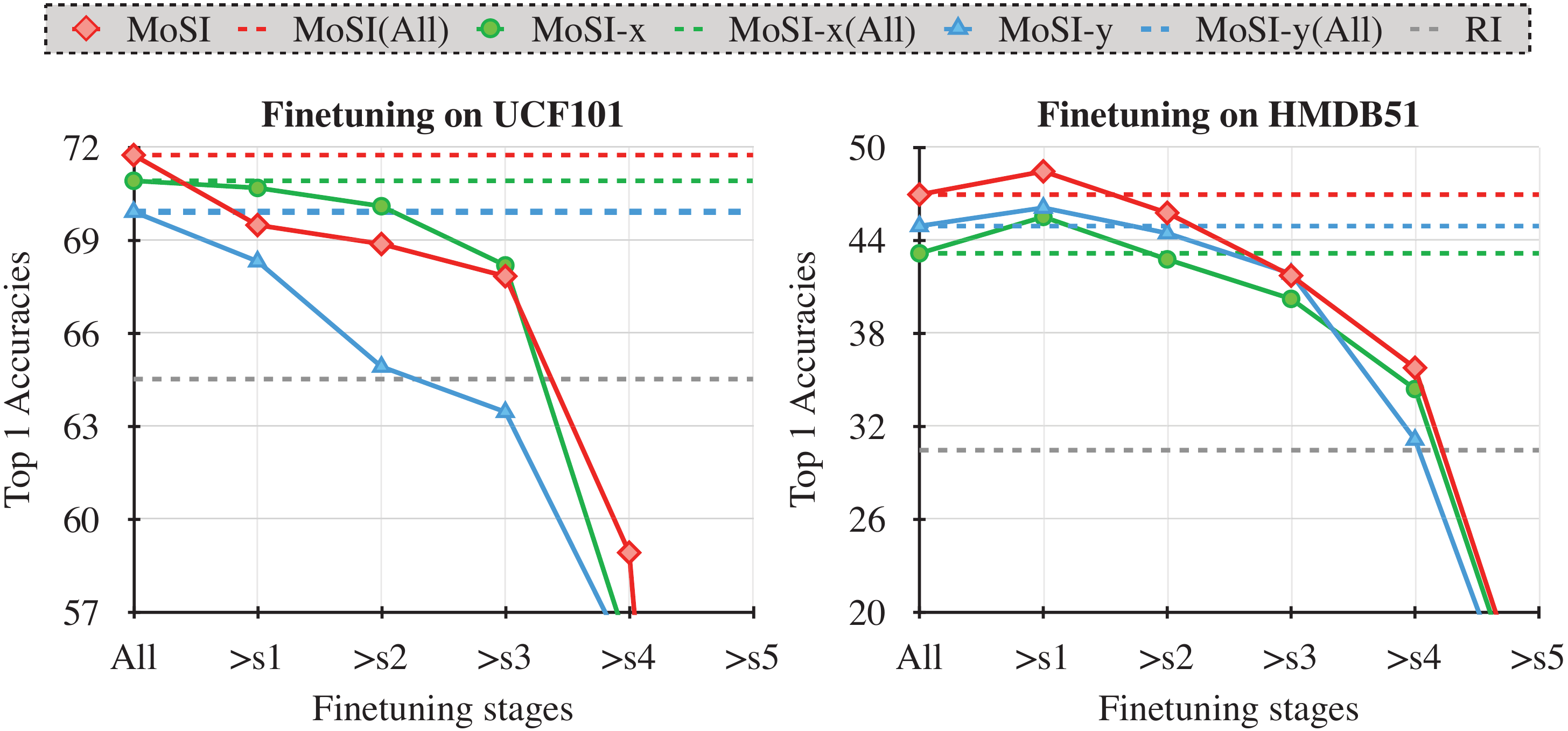}
\end{center}\vspace{-5mm}
   \caption{\textbf{Performance of the model trained by MoSI when the representation is frozen to a certain extent.} RI indicates training from random initialization, which is the baseline. The trend indicates the usefulness of the low-level feature learnt by MoSI. MoSI-x and -y indicates MoSI with label pool of respectively only horizontal and vertical labels. }
\label{fig:finetune}
\end{figure}

\noindent\textbf{Which parts of the representations learned using MoSI are the most useful?} 
We further fine-tune the pre-trained weights with different stages of the learned representation frozen, as in Fig.~\ref{fig:finetune}. 
By gradually freezing the representations during fine-tuning, we observe only a small drop in the performance for the first three stages. On HMDB51, fixing one stage even improves the accuracy. 
This indicate that the models can learn a strong low-level representation from MoSI.
Because MoSI focus less on the semantic understanding of the videos, only fine-tuning the linear layer does not have a high accuracy, which shows that the learned high-level representations are less discriminative. 
It is natural since the main objective of MoSI is for the network to attend to motions during fine-tuning.
The only information that the model receives is different motion patterns, while to discriminate between actions, not only the motion pattern, but also the identity of the moving object need to be recognized. 
Nevertheless, fine-tuning the last stage on HMDB still gives an improvement of $\sim 5\%$ over its random initialized baseline.
Comparing between MoSI with different label pools, we also observe a pattern similar to Table~\ref{tab:baseline}: On HMDB51, the models trained using only one axis consistently underperform the two-axes MoSI, while on UCF101, there is not a clear benefit of using two axes.
\begin{figure}[t]
\begin{center}
   \includegraphics[width=0.98\linewidth]{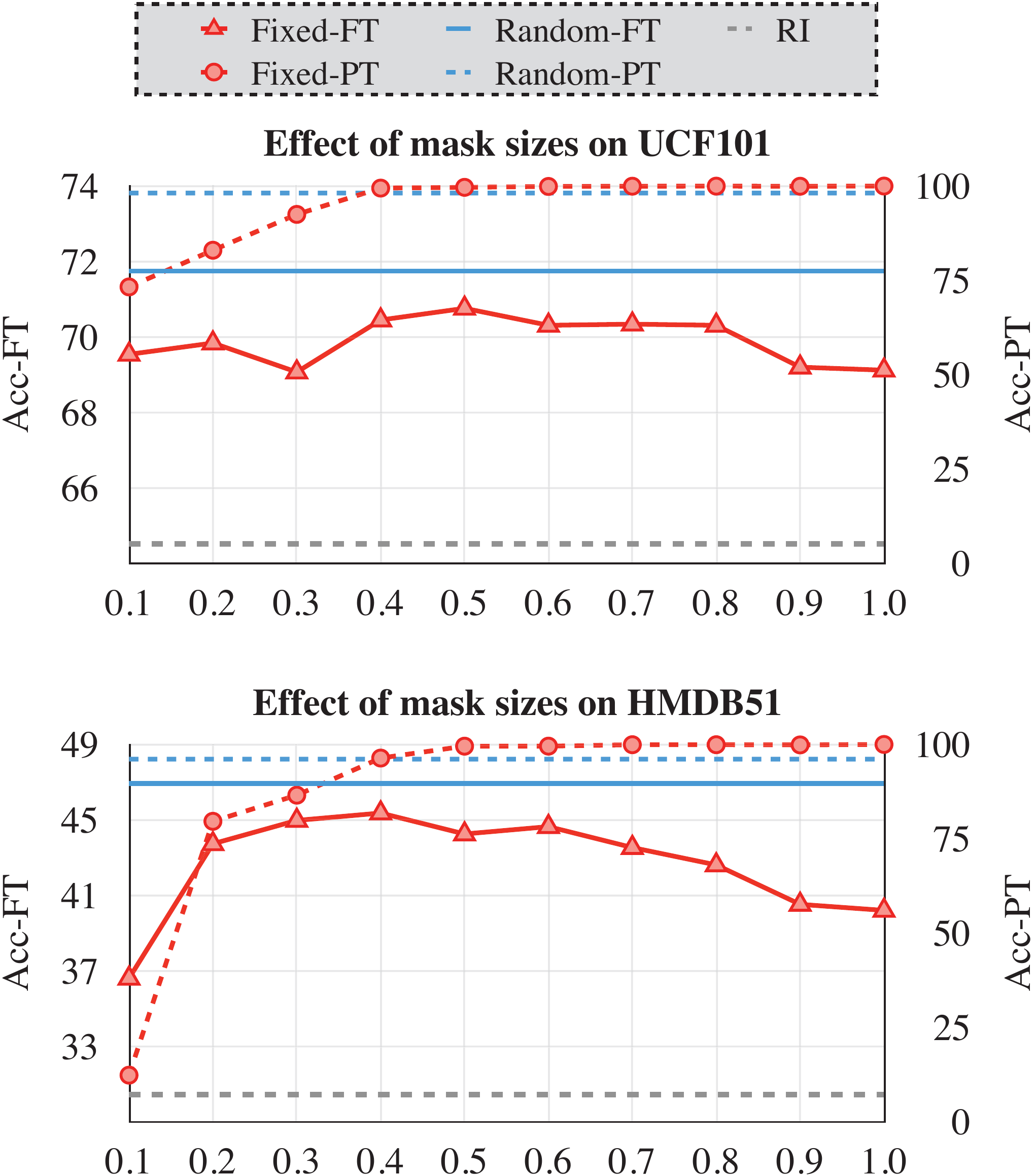}
\end{center}\vspace{-6mm}
   \caption{\textbf{Effect of mask sizes for MoSI.} The x-axis denotes the side length ratio $L_m/L$ of the unmasked area (1.0 as no static masks). Static mask is useful especially when the unmasked area covers a relatively small region. Random mask sizes further improve the recognition performance. }
\label{fig:mask_sizes}
\end{figure}

\subsection{Ablation Studies}
\noindent\textbf{Effects of mask sizes on MoSI.} 
We then investigate the effects of different mask sizes by altering the side length of the unmasked area $L_m$. 
The results are visualized in Fig.~\ref{fig:mask_sizes}. 
In terms of the pre-training accuracy, we can see that the MoSI task is generally easy on UCF101, with the pre-training accuracy being 73\% when $L_m/L$ is 0.1 and reaching 100\% for 0.4. 
This is probably because of the numerous similar visual contents in UCF101 that cause the network to memorize the patterns.
On the other hand, visual contents in HMDB51 have a larger diversity, thus the pre-training accuracy is lower compared to pre-training on UCF101 with the same parameters. 
For mask size ratio 0.1, the model can hardly learn to discriminate between different motions when the mask size $L_m=0.1\times L_s$.
This demonstrates that the proposed MoSI is not a trivial task that can be learned easily. 
Comparing different mask sizes, the validation accuracy during pre-training generally improves with the increase of the unmasked area. 
However, the high performance in our pretext task does not always mean a higher accuracy on the downstream task, which is also observed in \cite{revisiting}.
Therefore, a suitable mask size is crucial to ensure a high quality of the representations.
We then randomize the unmasked area within the range of $L_m=[0.3, 0.5]\times L_s$ and observe an improvement upon the variants with fixed mask sizes.

\noindent\textbf{Effects of the speed granularity and the number of frames on MoSI.}
As in Table~\ref{tab:hmdb51-classes} and \ref{tab:ucf101-classes}, compared to the default setting, reducing the number of class to $3$ hurt the performance in that the model is not able to distinguish different motion patterns, which shows the importance of the speed granularity. 
On the other hand, further increasing the granularity on top of 5 does not have a visible improvement as well. 
This means it is sufficient for the model to possess the basic ability to distinguish different speeds. 
For different frames, we fix the frame-wise distances for each label and alter the sizes of source images so that the only factor changed is the number of frames.
Table~\ref{tab:hmdb51-frames} and \ref{tab:ucf101-frames} show that using 16 frames to pre-train MoSI achieves the best performance. 
One possible reason is that the downstream task also uses 16 frames for fine-tuning. 

\begin{table}[t]\centering
\subfloat[\textbf{\# Classes on HMDB51}.  \label{tab:hmdb51-classes}]{
\tablestyle{3pt}{1.0}
\footnotesize
\begin{tabular}{c|cc}
 \small \# Class & \small Acc-PT & \small Acc-FT \\
\shline
 3 & 79.7 & 43.0 \\
 5 & 96.1 & \textbf{47.0}\\
 7 & \underline{96.3} & \underline{44.6} \\
 9 & \textbf{96.6} & 44.5 \\
\end{tabular}}\hspace{1mm}
\subfloat[\textbf{\# Classes on UCF101}. \label{tab:ucf101-classes}]{
\tablestyle{3pt}{1.0}
\footnotesize
\begin{tabular}{c|cc}
 \small \# Class & \small Acc-PT & \small Acc-FT \\
\shline
 3 & 81.4 & 70.1 \\
 5 & 98.2 & 71.8 \\
 7 & \underline{99.1} & 70.9 \\
 9 & \textbf{99.4} & 71.3 \\
\end{tabular}}\\\vspace{-1mm}

\subfloat[\textbf{\# Frames on HMDB51}. \label{tab:hmdb51-frames}]{
\tablestyle{3pt}{1.0}
\footnotesize
\begin{tabular}{c|cc}
 \small \# Frames & \small Acc-PT & \small Acc-FT \\
 \shline
 8   & 93.7 & \underline{45.6} \\
 12  & \underline{95.0} & 44.9 \\
 16  & \textbf{96.1} & \textbf{47.0} \\
 24  & 94.5 & 44.6 \\
 32  & 91.7 & 45.5 \\
\end{tabular}}\hspace{1mm}
\subfloat[\textbf{\# Frames on UCF101}. \label{tab:ucf101-frames}]{
\tablestyle{3pt}{1.0}
\footnotesize
\begin{tabular}{c|cc}
 \small \# Frames & \small Acc-PT & \small Acc-FT \\
 \shline
 8   & 98.1 & 71.7 \\
 12  & 98.3 & 71.0 \\
 16  & 98.2 & \textbf{71.8} \\
 24  & 98.1 & 71.1 \\
 32  & \textbf{99.6} & 70.0 \\
\end{tabular}}\\\vspace{-1mm}
\caption{\textbf{Ablation studies} on action recognition task with models trained by MoSI. Acc-PT and -FT denotes pre-training and fine-tuning top-1 accuracies respectively. For \# Classes, all parameters are kept the same except for the number of classes. For \# Frames, only the source image size changes with the number of frames to keep the motion magnitude unchanged. Bold and underlined numbers denotes the best and the second-best performance.
\label{tab:ablation}}
\end{table}
\begin{table}[]
    \centering
\tablestyle{3pt}{1.0}
\footnotesize
\begin{tabular}{c|cccccc|c}
 \small \# Samples & \small 3\scriptsize{(4\%)} & \small 5\scriptsize{(7\%)} & \small 7\scriptsize{(10\%)} & \small 9\scriptsize{(13\%)} & \small 11\scriptsize{(16\%)} & 13\scriptsize{(19\%)} & \small Full\\
\shline
 \scriptsize BASELINE & 4.5 & 6.7 & 8.1 & 10.4 & 11.7 & 14.8 & 30.4\\
 MoSI & \textbf{8.1} & \textbf{12.5} & \textbf{17.4} & \textbf{22.2} & \textbf{24.1} & \textbf{25.4} & \textbf{46.9}\\
 \hline
 Diff & +3.6 & +5.8 & +9.3 & +11.8 & +12.4 & +10.6 & +16.5\\
\end{tabular}\\\vspace{-1mm}
    \caption{\textbf{Low-shot fine-tuning on HMDB51.} Top-1 accuracy is used for comparison with the baseline (trained from random initialization).}
    \label{tab:fewshot}
\end{table}

\noindent\textbf{Few/low shot fine-tuning. }We also evaluate MoSI under a few/low-shot setting, where we randomly sample 3, 5, 7, 9, 11 and 13 videos from each class of the split 1 training set of HMDB51 as the training set for fine-tuning. 
This corresponds to using only around 4\% to 20\% of the original dataset. 
The results are demonstrated in Table~\ref{tab:fewshot}. 
Pre-trained using MoSI, the model is consistently better than the random initialized counterpart by a large margin. 
This shows the effectiveness of MoSI on few-shot video classification.

\noindent\textbf{Training on ImageNet.}
Since MoSI is able to train video models from static images, we additionally use ImageNet as the data source for pre-training. 
We only use a small portion of the original ImageNet data because the motion patterns can already be well learned on HMDB51 using only 5k videos. 
As in Table~\ref{tab:imagenet}, we further validate that the performance in the pretext task largely depends on the number of data.
Increasing the training data results in a higher validation accuracy before it saturates. 
In terms of the downstream task, the fine-tuning performance generally increases when the number of pre-training data increases before it saturates. 
Overall, the models pre-trained on ImageNet outperforms the ones trained on respective datasets. 

\begin{table}[t]
    \centering
\tablestyle{8pt}{1.0}
\footnotesize
\begin{tabular}{c|c|c|c}
 \small Dataset-PT & \small Acc-PT & \small Dataset-FT & \small Acc-FT \\
\shline
 UCF101 & 98.2 & \multirow{6}{*}{\small UCF101} & 71.8\\
 \cline{1-2}\cline{4-4}
 ImageNet-S5 & 83.1 &  ~& 71.1 \\
 ImageNet-S10 & 87.8 & ~ & 70.5 \\
 ImageNet-S20 & 96.9 & ~ & 71.2 \\
 ImageNet-S30 & 97.7 & ~ & \textbf{71.9} \\
 \hline
 HMDB51         & 96.1     & \multirow{5}{*}{\small HMDB51} & 47.0\\
 \cline{1-2}\cline{4-4}
 ImageNet-S5    & 83.1     & ~ & 47.3 \\
 ImageNet-S10   & 87.8     & ~ & 47.8 \\
 ImageNet-S20   & 96.9     & ~ & \textbf{48.0} \\
 ImageNet-S30   & 97.8     & ~ & 47.6 \\
\end{tabular}\\\vspace{-1mm}
    \caption{\textbf{Pre-training models using MoSI on ImageNet.} S5, 10, 20, 30 denote randomly sample 5, 10, 20, 30 from each class respectively. Training video models on ImageNet with MoSI further increases the recognition accuracy on downstream datasets.}
    \label{tab:imagenet}
\end{table}
\begin{table}[t]
    \centering
\tablestyle{4pt}{1.0}
\footnotesize
\begin{tabular}{ccc|cc}
 \multicolumn{3}{c|}{\small Initialization} & \multicolumn{2}{c}{Supervised fine-tuning}\\
 \small Method &\small Arch. &\small Dataset &\small UCF101 &\small HMDB51\\
 \shline
 OPN~\cite{opn} & VGG & UCF & 59.6 & 23.8 \\
 DPC~\cite{dpc} & R-2D3D & K400 & 75.7 & 35.7 \\
 MemDPC~\cite{memdpc} & R-2D3D & K400 & 78.1 & 41.2 \\
 3D-RotNet~\cite{3drotnet} & R3D & K400 & 62.9 & 33.7 \\
 ST-Puzzle~\cite{ST-puzzle} & R3D & K400 & 65.8 & 33.7\\
 VCP~\cite{vcp} & C3D & UCF/HMDB & 68.5 & 32.5 \\
 VCOP~\cite{vcop} & R(2+1)D & UCF & 72.4 & 30.9\\
 PRP~\cite{prp} & R(2+1)D & K400 & 72.1 & 35.0\\
 SpeedNet~\cite{speednet} & S3D-G & K400 & 81.1& 48.8\\
 \hline
 \textbf{MoSI (Ours)} & R-2D3D & UCF/HMDB & 71.8 & 47.0 \\
 \textbf{MoSI (Ours)} & R-2D3D & K400 & 70.7 & 48.6 \\
 \hline
 \textbf{MoSI (Ours)} & R(2+1)D & UCF/HMDB & \textbf{82.8} & \textbf{51.8} \\
\end{tabular}\\\vspace{-1mm}
    \caption{\textbf{State-of-the-art comparisons.} 
    }\vspace{-3mm}
    \label{tab:sota}
\end{table}
\subsection{Comparison with video-based methods}
In Table~\ref{tab:sota}, we demonstrate the performance comparison with the state-of-the-art video self-supervised training methods that only use RGB modality.
Overall, MoSI performs competitively against other methods on both UCF101 and HMDB51. 
Compared to DPC~\cite{dpc} and MemDPC~\cite{memdpc} with the same architecture MoSI achieves a much stronger performance on HMDB51. 
Note that DPC and MemDPC uses a 34-layer R-2D3D model with $224\times 224$ as input, while MoSI uses an 18-layer R-2D3D with $112\times 112$ as input. 
Using a stronger backbone R(2+1)D, we achieve the state-of-the-art performance on both UCF101 and HMDB51 datasets. 

\subsection{Discussions}
\begin{figure}[t]
\begin{center}
   \includegraphics[width=1\linewidth]{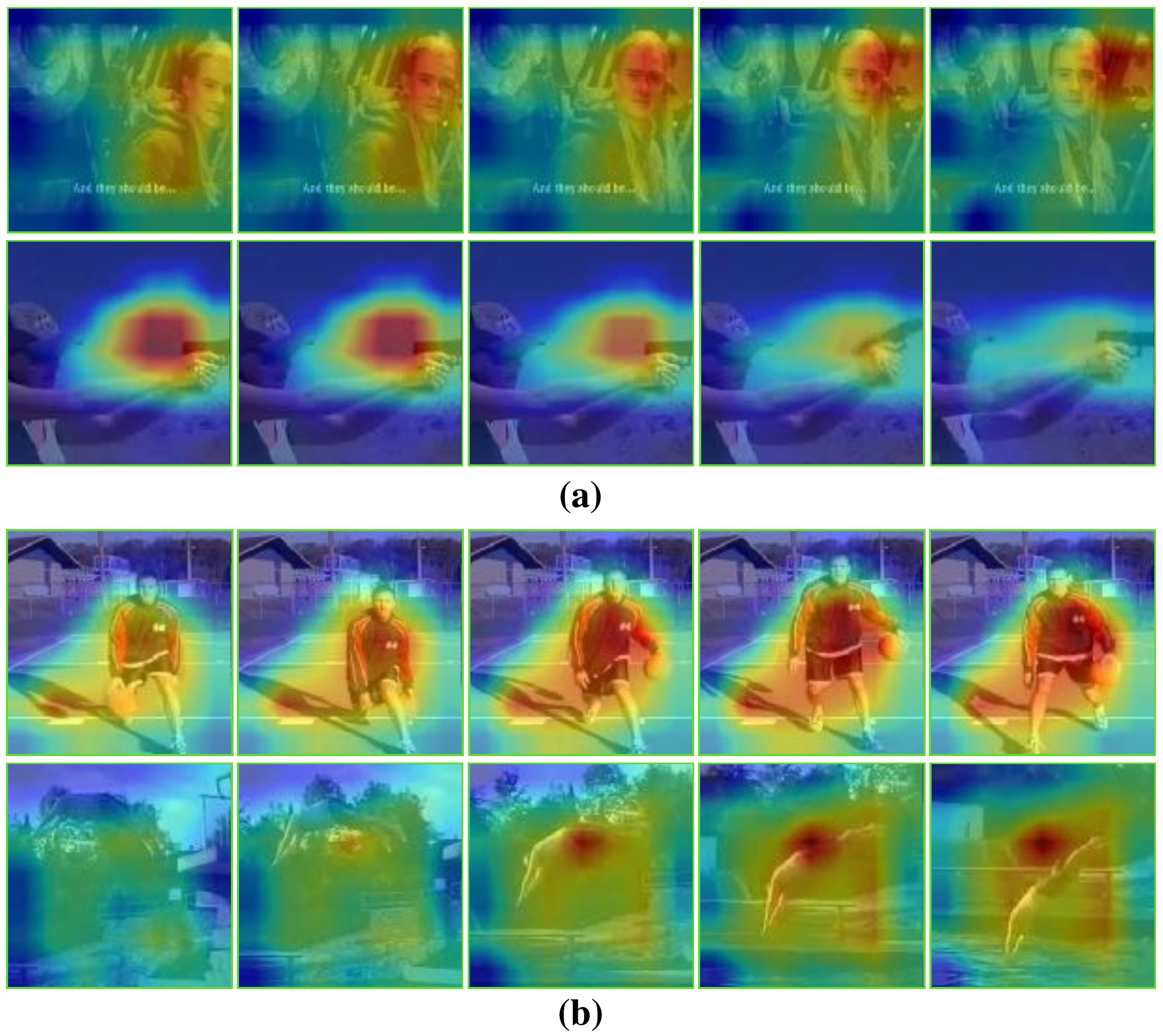}
\end{center}\vspace{-5.5mm}
   \caption{\textbf{Failure cases. }(a) Square activation patterns originated from the prior encoded by MoSI that the motion areas have a square shape. (b) The movement in the background causes confusion for the network. Examples show the model is confused by the movement of the shadow in the background and the scene respectively.}\vspace{-5mm}
\label{fig:failure_cases}
\end{figure}

Previous sections have shown that the proposed MoSI framework can train the model to focus on a local area with prominent motions. 
Despite that the models trained by MoSI achieves a satisfactory improvement on the action recognition task by learning to attend to motions, there are certain limitations. 
(A) Because of the square shape of the prior unmasked motion area in MoSI, in some cases, the model is somewhat biased toward attending to a square area as visualized in Fig.~\ref{fig:failure_cases}(a).
(B) Because no semantic information is encoded during MoSI training (see Sec.~\ref{sec:understanding}), major movements in the background could also confuse the model, as in Fig.~\ref{fig:failure_cases}(b). 

\noindent\textbf{Conclusion.}
This work proposes MoSI, a simple framework for the video models to learn motion representations from images. 
It is shown that MoSI can discover and attend to prominent motions in videos, thus yielding a strong representation for the downstream action recognition task.
We also demonstrate the possibility of using MoSI to train video models on image datasets. 
It is hoped that this research can inspire further study in understanding how motions can be encoded into video representations.

\noindent\textbf{Acknowledgement.} 
This research is supported by the Agency for Science, Technology and Research (A*STAR) under its AME Programmatic Funding Scheme (Project \#A18A2b0046) and by Alibaba Group through Alibaba Research Intern Program.

\begin{appendices}
\section{Appendix}
\noindent\textbf{Further training details. } 
During the self-supervised pre-training, we use Adam~\cite{adam} as our optimizer and use the learning rate of 0.001. The batch size is 10 source images per GPU for 16 GPUs. 
We use warm-up~\cite{warmup} of 10 epochs starting with a learning rate of 0.0001 and the total number of epochs is 100 except for kinetics, where 2 warm-up epochs and in total 20 epochs are used to train the model. A half-period cosine schedule~\cite{cosinelr} is adopted. A dropout of 0.5 is used during the pre-training of the video models. Besides the MoSI-specific augmentations, including random sizes and locations of the static mask, we also apply a frame-wise random color jittering following~\cite{dpc,memdpc}, for the model to learn not only low-level pixel-correspondence, but also semantic correspondences. 
The weight decay is set to 1e-4.
During fine-tuning on the downstream action classification task, warm-up is applied with a starting learning rate of 1/10 the base learning rate. The weight decay is set to 0.001. Ten warmup epochs is used and the model is trained in total for 300 epochs. Data augmentation include spatial random crop, random horizontal flip, and clip-wise random color jittering with the same parameter as in self-supervised training. 
  \end{appendices}

{\small
\bibliographystyle{ieee_fullname}
\bibliography{egbib}
}

\end{document}